\documentclass[11pt]{article}
\usepackage{arxiv}   
\usepackage{booktabs}
\usepackage{amsmath}
\usepackage{graphicx}
\usepackage{url}
\usepackage[numbers]{natbib}

\title{Can a Model Catch Its Own Hallucinations for Free?: Label-Free Doubt Signals
Hold Their Own Against a Labelled Dataset for Abstention}

\author{%
  Ali Asaria \\ Transformer Lab \and
  Tony Salomone \\ Transformer Lab \and
  Deep Gandhi\thanks{Corresponding author: \texttt{deep@lab.cloud}} \\ Transformer Lab
}
\date{}
\runningtitle{Label-Free Calibration Fine-Tuning for Hallucination Reduction}

\begin{document}
\maketitle

\begin{abstract}
Large language models state false facts as fluently as true ones, yet a model often ``knows''
internally when it is on shaky ground: the probability it assigns to its own answer tends to dip
on the facts it gets wrong. The usual way to act on this, teaching a model to abstain rather than
guess, requires a labelled dataset of right and wrong answers. We ask whether the model's
\emph{own} confidence, which is free and needs no labels, can do that job instead. We fine-tune
each model (with LoRA) to answer when its frozen confidence is high and to say ``I'm not sure''
when it is low, using the signal alone and no correctness labels. Across six open-weights models
(1B--8B, two families) on short-form factual question answering, with correctness adjudicated by
an independent judge model, this label-free recipe holds its own against label-supervised
abstention-tuning: at matched coverage we find no statistically detectable difference between the
two. A control that drills hard examples instead of abstaining does not help, indicating the gain
comes from \emph{calibration}, not rote memorization. The signal's one blind spot is
\emph{confidently wrong} facts, which it cannot flag. A model's own doubt is thus a near-free
substitute for a labelled dataset when teaching it when to abstain. Code and artifacts are
available on request.
\end{abstract}

\section{Introduction}
Language models hallucinate: they assert false facts as confidently as true ones, and a fluent
surface gives a reader no signal that a claim is unsupported. A growing body of work observes
that a model's \emph{internal} state nonetheless carries information about its own
correctness, in particular that token-level probabilities are higher on facts the model gets
right than on facts it gets wrong. We confirm this on our own models: the mean log-probability
over an answer span separates correct from incorrect answers with AUROC $0.73$--$0.86$, holding
even for a 1B model.

If a model already ``knows when it is unsure,'' the natural question is whether it can be taught
to \emph{act} on that knowledge, to abstain on its low-confidence facts instead of asserting
them. One could simply threshold the frozen signal at inference time as an external gate; we
instead \emph{internalize} the behavior into the weights so the model abstains natively, which
needs no external scorer at deployment and lets abstention interact with generation. The usual
way to teach abstention is supervised: collect correctness labels for a training set and train
the model to decline the questions it gets wrong. But labels are expensive: they require knowing
the right answer to every training question. We ask a cheaper question: can the model's
\emph{own confidence signal} (which is free, already present, and requires no labels) play the
role of those labels?

We study \emph{signal-gated abstention fine-tuning}. We compute, once, the frozen base-model
confidence over each training question's answer span; threshold it; and fine-tune the model so
that low-confidence questions are retargeted to ``I'm not sure'' while high-confidence questions
keep the model's own answer. No correctness label ever enters this procedure. We compare it
head-to-head against the label-supervised version of the same recipe and against standard
fine-tuning, an up-weighting control, and the untrained base, on a $2\times3$ model grid, with
correctness adjudicated by an independent open-weight judge and significance assessed by
bootstrap confidence intervals.

\textbf{Contributions.}
\begin{enumerate}
  \item \textbf{A label-free recipe for abstention.} We introduce \emph{signal-gated abstention
    fine-tuning}, which uses a model's own frozen token-probability confidence, with no
    correctness labels, to teach it to abstain on the facts it is internally unsure of. We first
    verify the premise the recipe rests on: this confidence discriminates the model's own
    hallucinations across six models and two families, even at 1B scale (\S\ref{sec:results}).
  \item \textbf{Free confidence holds its own against a labelled dataset.} At matched coverage and
    adjudicated by an independent judge, the label-free recipe is competitive with
    label-supervised abstention-tuning: we detect no difference on any of six models
    (\S\ref{sec:results}).
  \item \textbf{The gain is calibration, not memorization.} An up-weighting control that drills
    the same hard questions without abstaining does not reduce hallucination, isolating
    calibration as the mechanism (\S\ref{sec:results}).
  \item \textbf{An honest account of the limits.} We map where the method fails: confidently-wrong
    facts the signal cannot flag, and the rare entities where hallucination concentrates
    (\S\ref{sec:discussion}).
\end{enumerate}

\section{Related Work}
\paragraph{Internal confidence signals.} A line of work reads a model's own internal state to
detect hallucination: token-probability-based hallucination detectors \citep{2512.03107v1,
2504.07863v3}, and surveys of factual-confidence estimators that caution that raw sequence
probability is a comparatively weak estimator on QA \citep{2406.13415v1}. \citet{2604.22271v2}
argue that first-order token log-probabilities have intrinsic blind spots and propose verbalized
or activation-based alternatives. Our results are consistent with both observations: the
log-probability signal is discriminative enough to be useful, yet its residual failures are
exactly the confidently-wrong cases it cannot see.

\paragraph{Teaching abstention and refusal.} A complementary line fine-tunes models to abstain
at their knowledge boundary, typically using correctness signals, labels, or rewards:
refusal-aware tuning and refusal tokens \citep{2412.06748v2}, knowledge-boundary abstention via
reinforcement learning \citep{2604.22779v1}, and training-free conformal abstention
\citep{2604.27914v1}. These methods rely on correctness signals, labels, or rewards to decide
\emph{where} to abstain; we ask whether the model's own confidence can supply that supervision
for free.

\paragraph{Internalizing confidence by training.} Several methods bake a confidence or abstention
behavior into the weights: confidence tokens and routing \citep{2410.13284v3}, confidence tuning
for cascades \citep{2502.19335v3}, differentiable calibration losses \citep{2412.02904v2}, and
reinforcement learning with proper-scoring rewards \citep{2512.19920v3}. The closest to ours are
self-supervised: distilling a model's own (sampling- or self-evaluation-based) uncertainty into
verbalized confidence \citep{2503.14749v3, 2409.12180v1} and self-distillation that restores
calibration with little external labeling \citep{2603.06604v1}. We differ in the specific, cheap
recipe: a \emph{frozen, single-pass token-probability} signal thresholded into a binary
abstain/answer target, and in a controlled, coverage-matched comparison to the
\emph{label-supervised} version of the same recipe. \citet{2604.15574v1} document that standard
fine-tuning can induce factual forgetting; our standard-fine-tuning control reproduces this
neutral-to-slightly-worse behavior. Finally, \citet{2505.24858v2} caution that calibration
metrics and faithful uncertainty expression can diverge; we therefore report selective risk
rather than relying on expected calibration error.

\section{Method}
\label{sec:method}
\paragraph{The confidence signal.} For a question $q$, the model greedily decodes an answer
$a=(t_1,\dots,t_m)$. We define its confidence as the mean log-probability the model assigns to
its own answer tokens,
\[
s(q) \;=\; \frac{1}{m}\sum_{i=1}^{m}\log p_\theta\!\left(t_i \mid q, t_{<i}\right),
\]
computed \emph{once} from the frozen base model before any fine-tuning.
Higher $s$ means the model concentrated probability on its answer; lower $s$ means it hedged
internally even while producing a fluent answer.

\paragraph{Signal-gated abstention fine-tuning (ours, ``C3'').} On a training set we rank
questions by $s(q)$ and pick the lowest-$s$ fraction (the abstention threshold, a coverage
knob). For those low-confidence questions the supervised target becomes a fixed abstention
string (``I'm not sure.''); for the rest the target is the model's \emph{own} greedy answer
(self-distillation). We fine-tune with LoRA ($r{=}16$). \textbf{No correctness label is used}:
the frozen signal alone decides where to abstain.

\paragraph{Comparators (same recipe, different supervision).}
\begin{itemize}
  \item \textbf{Base}: the untrained instruction model.
  \item \textbf{Standard SFT (``C1'')}: LoRA on the model's own greedy answers everywhere (no
    abstention, no signal).
  \item \textbf{Label-supervised abstention (``C2'', R-Tuning style \citep{rtuning})}: identical to ours except
    the abstain/answer decision uses the \emph{correctness label} (abstain where the base model
    was wrong, by gold+alias match) instead of the signal.
  \item \textbf{Up-weighting control (``C4'')}: instead of abstaining on low-signal questions,
    \emph{up-weight} their loss toward the answer (``memorize harder''). If C4 matched C3, the
    gain would be recall, not calibration.
\end{itemize}

\section{Experimental Setup}
\label{sec:setup}
\paragraph{Models.} Six open-weights instruction models forming a $2\times3$ grid (family
$\times$ scale): Llama-3.2-1B, Llama-3.2-3B, Llama-3.1-8B \citep{llama3}; Qwen3-1.7B/4B/8B
\citep{qwen3} (run non-thinking for answer-span comparability). All fine-tuned with LoRA
\citep{lora2021} ($r{=}16$); one training run per cell.

\paragraph{Data.} A factual short-question-answering set of 700 development + 700 test questions
drawn from PopQA \citep{popqa} (MIT) and TriviaQA \citep{triviaqa} (Apache-2.0), 300 TriviaQA +
400 PopQA per split.

Splits are \emph{entity-grouped}: development and test share \emph{zero} entities, preventing
memorized-fact leakage.
On-policy answers and the confidence signal are generated per model.
\paragraph{Metrics and protocol.} \emph{Coverage} is the fraction of questions a model answers
(rather than abstaining); \emph{hallucination rate} is the fraction of \emph{answered} items the
judge marks incorrect. Because methods that abstain more answer fewer (and easier) questions,
the two trade off: the \emph{risk--coverage curve} plots hallucination rate as the abstention
threshold sweeps coverage, and AURC (area under it, lower is better) summarizes selective
prediction independent of any single operating point. Correctness on the held-out test set is
adjudicated by an \emph{independent} open-weight judge, \texttt{gemma-2-27b-it} \citep{gemma2} (deliberately
neither the Llama nor the Qwen family under test, to avoid same-family self-favoring).
We use AURC rather than expected calibration error because our models abstain rather than emit a
numeric confidence, so a confidence-bin calibration metric is ill-posed here.
\paragraph{Significance.} \emph{Matched coverage} means we set C3's abstention threshold per
model so its coverage approximates C2's, then compare hallucination at that operating point
(small residual coverage gaps remain; see Limitations). Confidence intervals are percentile
intervals from 1{,}000 item-level bootstrap resamples of the 700-item test set (fixed seed).
 These intervals capture test-set sampling variance only, not training-seed variance.

\section{Results}
\label{sec:results}

\paragraph{The signal discriminates hallucination (and largely survives scale).}
The frozen mean-log-probability signal separates correct from incorrect base-model answers with
AUROC $0.778/0.821/0.858$ (Llama 1B/3B/8B) and $0.755/0.778/0.725$ (Qwen3 1.7B/4B/8B), all
above $0.65$, and discriminative even at 1B.
The trend is monotone within Llama but not within Qwen3 (its 8B value is the lowest of the
three).

\paragraph{Label-free is competitive with label-supervised (headline).}
Table~\ref{tab:main} reports judge-adjudicated hallucination at matched coverage. On every model
the C2-vs-C3 95\% bootstrap confidence intervals overlap, i.e.\ we do not detect a hallucination
difference between the label-free and label-supervised methods.
Point estimates fall within a narrow band (C3/C2 reduction ratio $0.95$--$1.10$) that is, on this
evidence, within noise; we therefore read the result as parity (failure to detect a difference),
not as either method beating the other. Standard fine-tuning (C1) is neutral-to-slightly-worse
than base.

\begin{table}[t]
\centering
\caption{Judge-adjudicated hallucination rate (lower is better) on the 700-item held-out test
set. \textbf{C1}=standard SFT, \textbf{C2}=label-supervised abstention (R-Tuning), \textbf{C3}=
ours (label-free). C3 is evaluated at coverage matched to C2; C3 coverage shown (C2 operates at
comparable low coverage, $\approx$0.15--0.50, by over-abstaining). On every model the C2-vs-C3
95\% bootstrap CIs overlap (no detected difference). The C3/C2 column (ratio of hallucination
\emph{reduction} vs.\ base) is a point estimate within noise.}
\label{tab:main}
\begin{tabular}{lccccc}
\toprule
Model & Base & C1 & C2 (labels) & C3 (ours; coverage) & C3/C2 \\
\midrule
Llama-3.2-1B & 0.707 & 0.740 & 0.400 & 0.385 \,(0.19) & 1.05 \\
Llama-3.2-3B & 0.594 & 0.553 & 0.211 & 0.171 \,(0.28) & 1.10 \\
Llama-3.1-8B & 0.445 & 0.461 & 0.149 & 0.164 \,(0.38) & 0.95 \\
Qwen3-1.7B   & 0.759 & 0.750 & 0.316 & 0.286 \,(0.08) & 1.07 \\
Qwen3-4B     & 0.660 & 0.661 & 0.218 & 0.185 \,(0.09) & 1.07 \\
Qwen3-8B     & 0.579 & 0.580 & 0.208 & 0.179 \,(0.19) & 1.08 \\
\bottomrule
\end{tabular}
\end{table}

\begin{figure}[t]
\centering
\includegraphics[width=0.85\linewidth]{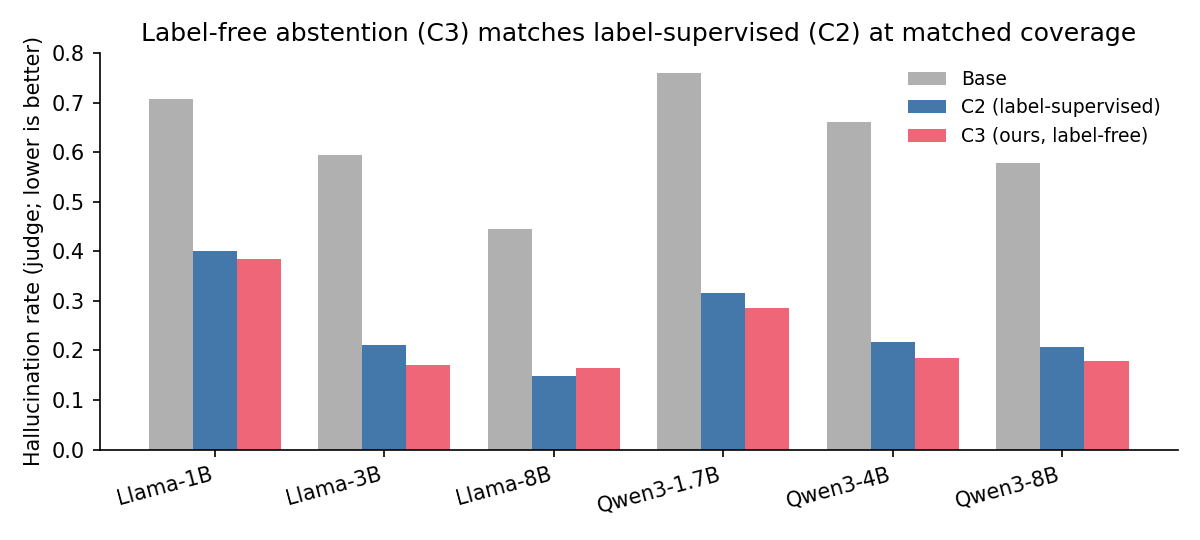}
\caption{Hallucination rate (lower is better). At matched coverage, label-free abstention
(C3, red) and label-supervised abstention (C2, blue) show no detected difference on any model
(overlapping 95\% bootstrap CIs), both far below the untrained base (grey). The label-free
method needs no correctness labels.}

\label{fig:main}
\end{figure}

\paragraph{The gain is consistent with calibration, not memorization.}
The up-weighting control (C4) stays at base hallucination with near-full coverage, whereas C3
reduces hallucination: drilling low-signal questions harder does nothing, while retargeting them
to abstention helps.
(This control is measured at our string-match, $n{=}250$ stage; the selective-prediction result
below, judge-based at $n{=}700$, is stronger evidence for genuine calibration.)

\paragraph{Selective prediction.}
Treated as a selective predictor, C3's frozen signal yields lower area under the risk--coverage
curve than a non-selective model: $7/25/28\%$ for Llama (1B/3B/8B) and $17/24/26\%$ for Qwen3
(1.7B/4B/8B), with the largest gains for the larger Llama models.
Abstention is well-targeted: roughly a 7:1 ratio of useful corrections to abstentions wasted on
correct answers.

\paragraph{Robustness.}
The parity conclusion is unchanged under string-match vs.\ judge correctness (the judge makes C3
look slightly \emph{better}, not worse), holds across both model families, and is stable from
$n{=}250$ to $n{=}700$; indeed an apparent $69\%$ point estimate at $n{=}250$ was sampling noise
that vanished under the $n{=}700$ bootstrap.
Substituting an energy signal for mean log-probability gave no gain.

\section{Discussion and Limitations}
\label{sec:discussion}
The practical takeaway is that a model's own confidence is a promising near-free substitute for
correctness labels when teaching short-form abstention: across six models we could not
distinguish the label-free method from the label-supervised one.
Two failure modes bound the result. First, hallucination concentrates on \emph{rare}
entities: the worst popularity quartile runs roughly 2--3$\times$ the best (e.g.\ Llama-8B
$0.89$ vs.\ $0.32$); the method handles these by abstaining rather than by knowing rare facts.

Second, and dominant, are \emph{confidently wrong} answers: $36\%$ of all items are
answered-but-wrong with confidence high enough that the signal does not flag them, the bulk of
the $39\%$ answered-but-wrong residual.
These are high-signal items on which C3 keeps (and thus self-distills) the model's own wrong
answer; our calibration claim is therefore specifically about the \emph{abstention} behavior on
low-signal items, not that the method is memorization-free overall. This is the intrinsic ceiling
of a single-pass log-probability signal and motivates richer signals (verbalized confidence,
activation probes) as future work.

\subsection{Limitations}
A few refinements would further strengthen these results. Our parity finding is a failure to
detect a difference rather than a formal equivalence, so a paired equivalence test (TOST) at
exactly-pinned coverage is a natural next step, alongside a multiple-comparison adjustment across
the six models. The two methods are compared at approximately matched coverage, with the small
residual gap if anything favouring the label-free method, so we report parity rather than
superiority. Stronger baselines are also worth exploring, such as a judge-trained label-supervised
method and a second labelled recipe, and the up-weighting control could be extended to the full
evaluation set. Finally, the study covers English short-form question answering at modest scale,
so extending the approach to long-form generation and larger models is promising future work;
correctness throughout is decided by a single judge model, with a string-match measure reported
alongside as a check.

\section{Availability}
The training and evaluation code, the fine-tuned LoRA adapters, and the data-construction scripts
are written and available on request from the corresponding author. No artifacts are
hosted publicly at this time.

\section{Conclusion and Future Work}
On English short-form factual QA at $\leq$8B, a model's own token-probability confidence, frozen
and used without any correctness labels, can be fine-tuned into native abstention that is
competitive with label-supervised abstention-tuning across six open-weights models, at no
labeling cost; we do not detect a difference between the two. The gain is consistent with
calibration rather than memorization, and selective-prediction value tends to grow with scale
within the Llama family. The chief limitation is confidently-wrong facts the signal cannot see.
The most promising next steps are a formal equivalence test at exactly-pinned coverage, a
judge-trained and a second label-supervised baseline, richer self-supervised signals (verbalized
confidence, activation probes), and extension to long-form factuality.

\bibliographystyle{plainnat}
\bibliography{references}
\end{document}